\pdfoutput=1

\documentclass[11pt]{article}

\usepackage[]{acl}

\usepackage{times}
\usepackage{latexsym}

\usepackage[T1]{fontenc}

\usepackage[utf8]{inputenc}

\usepackage{microtype}

\usepackage{inconsolata}
\usepackage{booktabs}
\usepackage{tabularx, ragged2e}
\usepackage{xcolor}
\usepackage{graphicx}
\usepackage{caption}
\usepackage{hyperref}
\usepackage{url}
\usepackage{comment}
\usepackage{array}

\def\@fnsymbol#1{\ensuremath{\ifcase#1\or *\or \dagger\or \ddagger\or
   \mathsection\or \mathparagraph\or \|\or **\or \dagger\dagger
   \or \ddagger\ddagger \else\@ctrerr\fi}}

\title{Subtle Misogyny Detection and Mitigation: An Expert-Annotated Dataset}

  \author{
  Brooklyn Sheppard\textsuperscript{1,*},
   {\bf  Anna Richter\textsuperscript{1,*}},
  {\bf  Allison Cohen\textsuperscript{1,$\dagger$}},
  {\bf  Elizabeth Allyn Smith\textsuperscript{2}},\\
  {\bf  Tamara Kneese\textsuperscript{3}},
  {\bf  Carolyne Pelletier\textsuperscript{4}},
  {\bf  Ioana Baldini\textsuperscript{5}},
  {\bf  Yue Dong\textsuperscript{6}}\\
  \textsuperscript{1} Mila - Quebec AI Institute
  \textsuperscript{2}  Université du Québec à Montréal
  \textsuperscript{3} Data \& Society Research Institute\\
  \textsuperscript{4} Mantium
  \textsuperscript{5} IBM Research
  \textsuperscript{6} University of California, Riverside\\
  \texttt{\{anna.richter, brooklyn.sheppard, allison.cohen\}@mila.quebec}\\
  \texttt{smith.eallyn@uqam.ca, tkneese@datasociety.net} \\
  \texttt{carolyne.pelletier@mantiumai.com, ioana@us.ibm.com, yue.dong@ucr.edu}
}

\begin{document}
\maketitle
\begin{abstract}
Using novel approaches to dataset development, the Biasly dataset captures the nuance and subtlety of misogyny in ways that are unique within the literature. Built in collaboration with multi-disciplinary experts and annotators themselves, the dataset contains annotations of movie subtitles, capturing colloquial expressions of misogyny in North American film. The dataset can be used for a range of NLP tasks, including classification, severity score regression, and text generation for rewrites. In this paper, we discuss the methodology used, analyze the annotations obtained, and provide baselines using common NLP algorithms in the context of misogyny detection and mitigation. We hope this work will promote AI for social good in NLP for bias detection, explanation, and removal. 

\textcolor{red}{\textbf{Content Warning:} To illustrate examples from our dataset, misogynistic language is used in Section 2.1, Table 2, and Table 3, which may be offensive or upsetting.} 
\footnotetext{$*$ Equal contributing first authors, please cite as Sheppard \& Richter  et al.}
\footnotetext{$\dagger$ Corresponding author: allison.cohen@mila.quebec}
\end{abstract}

\section{Introduction and Related Work}

An important but overlooked factor of socially responsible language modelling (LM) research and deployment is the creation process and quality of the datasets that LMs are trained on \citep{stochasticparrots, datasheets}.
When using LMs to perform sensitive, subjective, and societally impactful tasks like misogyny detection, hate speech mitigation, or online content moderation, the quality of the underlying dataset is critical \citep{Accountability}.
Because the model will align to the biases in the dataset, which often reflect the biases, or oversights, of the dataset creators \citep{bias-in-hatespeech}, it is crucial to include a diverse group of stakeholders in the dataset creation process, including LM domain experts and stakeholders who would be impacted by any deployed model that was trained on the dataset \citep{Multidisciplinary, gavin}.

Dataset work in the field of bias and more specifically sexism or misogyny detection has mainly focused in recent years on the domains of social media, with data stemming from Twitter, Reddit or Gab \citep{Edos, Guest, Ami} (see Table \ref{tab:dataset_comparison}).
While using this type of training data is valuable for detecting the often blatant and strong forms of misogyny appearing in social media forums, we contend that those datasources might not be ideal for detecting subtler forms of misogyny found in everyday spoken language, as they might overshadow the latter during the training process \citep{reif2023fighting}. Studies with movie or sitcom subtitles as training data may represent a better balance; in this domain \citet{hollywood} focuses on the elimination of all types of bias, and \citet{sitcoms} does not provide sufficient detail to permit comparison.

\begin{table*}
\centering
\resizebox{\textwidth}{!}{%
\small

\begin{tabular}{p{2.4cm}|r|ccc|rrrr}
\toprule
Dataset & Size  & Classifi. & Severity & Mitigation & Annotators & StM & \# Annot. & Source\\
\midrule
EDOS (\citeyear{Edos}) & 20,000  &   Y  & -  & - & Trained annotators & Y & 19 & Reddit, Gab\\
Guest (\citeyear{Guest}) & 6,567  &   Y & - & - & Trained annotators & Y & 6 & Reddit \\
Ami (\citeyear{Ami}) & 5,000 &   Y   & - & - &   Domain experts& Y & 6 & Twitter\\
Callme (\citeyear{callme}) & 13,631 &   Y   & -  &  Y & Crowdworkers &  Y & -  & Twitter, Psych. Scales   \\
ParaDetox (\citeyear{logacheva-etal-2022-paradetox}) &  11,939      & - &- & Y & Crowdworkers & N&   & Twitter, Reddit, Jigsaw \\
APPDIA (\citeyear{atwell2022appdia}) & ~ 2,000  & -& - & Y & Domain experts & N & - & Reddit\\
\midrule
Biasly (ours) & 10,000& Y& Y& Y & Domain experts & Y & 10 & Movie subtitles\\
\bottomrule 
\end{tabular}
}
\caption{Comparison of misogyny detection and bias mitigation datasets. `StM' denotes specificity to Misogyny; `Classifi.' indicates support for classification tasks; `\# Annot.' refers to the number of annotators.}
\label{tab:dataset_comparison}
\end{table*}

Even though most datasets provide a more fine-grained classification for different subtypes of misogyny \citep{Edos, Guest, Ami, callme}, the detection of misogyny remains, at its core, a classification problem where a (sub)category can be either present or not.
We argue that due to its nuanced and subjective nature, a continuous severity score modelled by regression is better suited for the detection of subtle misogyny.
Only one misogyny-specific dataset presents a form of misogyny mitigation \citep{callme}.
Their goal was to create adversarial examples that language models would find hard to differentiate from real sexist statements, by applying minimal lexical changes.
Our work is methodologically closer to the ParaDetox \citep{logacheva-etal-2022-study} and APPDIA \citep{atwell2022appdia} datasets, which released a parallel corpus for detoxicification.
To our knowledge, our dataset is the first parallel corpus with the purpose of training language models to rewrite text to mitigate the subtle misogyny contained therein.

In this work, we document the creation process of the Biasly dataset, an expert annotated dataset for the detection and mitigation of subtle forms of misogyny.
We describe our process, including the way we thoughtfully select, train, and engage with our annotators, ensuring our dataset is both high quality and created in a socially responsible way.
Finally, we present a short analysis of the annotated dataset and provide model baseline results for the tasks of misogyny classification, severity prediction and mitigation.

\section{Dataset Creation}

We believe that an interdisciplinary team fosters responsible research in language modeling. For the focus of this project, the most relevant domains of expertise were gender studies and linguistics to ensure the annotated data is attuned to misogyny and the various, complex forms it takes in language. Our humanities and social science experts recruited expert annotators from both areas to help shape the project from its earliest stages, as detailed in section \ref{annotators}. To do this work responsibly, we made sure that these experts interacted with the data in its various forms at every step of the project. We describe some benefits that came from looking closely at the data in what follows. 

\subsection{Dataset choice}
\textbf{Contemporary movie subtitles}:
Biasly's data is derived from a movie subtitle corpus available through \textit{English-corpora.org}. While movie scripts themselves might seem preferable to automatically generated subtitles, this dataset was the only one we found with sufficient quantity for our task. The decision to use movie subtitles was motivated by 2 objectives: 1) the presence of both overt and subtle forms of misogyny in good proportion, and 2) its similarities to transcribed conversational speech. Because Twitter, Reddit, and Gab are known to offer an abundance of overt misogyny, it was a concern that these more overt forms would predominate and drown out the effect of subtle examples \citep{reif2023fighting}. Business e-mail corpora were rejected for a lack of misogynistic language in sufficient quantity for analysis. Movie subtitles, however, offered both types of misogyny in the necessary quantity and proportion. We sought to complement existing efforts that focus on written language with an analysis of spoken language because differences in communication type lead to differences in misogynistic expression. Though scripts are written and not naturalistic speech, screenwriters try to create fluid verbal interactions, and the subtitles from the films are gleaned from speech-to-text algorithms designed to capture oral communication.



\paragraph{Data Pre-processing:}
Biasly's humanities and social science experts analyzed a randomly-drawn sample of the data to determine necessary pre-processing techniques. Given how significantly language evolves over time \citep{juola2003time}, and how differently some items would be judged in one context versus another (e.g. \textit{lil darlin'}), we restricted our sample to movies released in the last 10 years. We filtered out films that, while contemporary productions, were set in the past (e.g. westerns, period pieces) or otherwise did not reflect contemporary colloquial speech (e.g. documentaries). Similarly, we reduced the sample to films that were American releases, given dialect differences across global Englishes \citep{major2005testing}. We also removed movies for which the subtitles were entirely upper- or lowercase to acknowledge the differences in meaning that this changing case produced (i.e. \textit{Black woman} versus \textit{black woman}, \textit{Karen} versus \textit{karen}, \textit{bitch} versus \textit{BITCH}). Furthermore, we filtered out explicitly-indicated speaker changes since this variable's inclusion was not constant across subtitles and would have affected the consistency of annotators' assumptions about the speakers and their intentions. 
Finally, we parsed the data into non-overlapping chunks of three sentences each, subsequently referenced as ``datapoints."


\paragraph{Data Filtration Approach:}
In order to identify as much misogyny as possible without biasing the dataset with terms that were already potentially misogynistic on their own (e.g. \textit{bitch} or feminine-specific job titles), we further filtered the data as follows: 
20 percent of our datapoints contain the keyword \textit{she}, 20 percent \textit{her}, 10 percent \textit{herself}, 10 percent \textit{women}, and 10 percent \textit{woman}. The remaining 30 percent were sampled randomly.
This data split roughly respects the bias of “natural" occurrences of these keywords in the dataset (\textit{she} and \textit{her} are used twice as often as the other keywords, reflecting their relative frequency in the overall corpus). Though these pronouns reflect a third-person orientation because \textit{you} and others are not gendered, the 30 percent random sample included directly-addressed misogyny as well as other types.

\subsection{Annotation Tasks/Taxonomy}
\begin{figure}[h]
\includegraphics[width=\columnwidth]{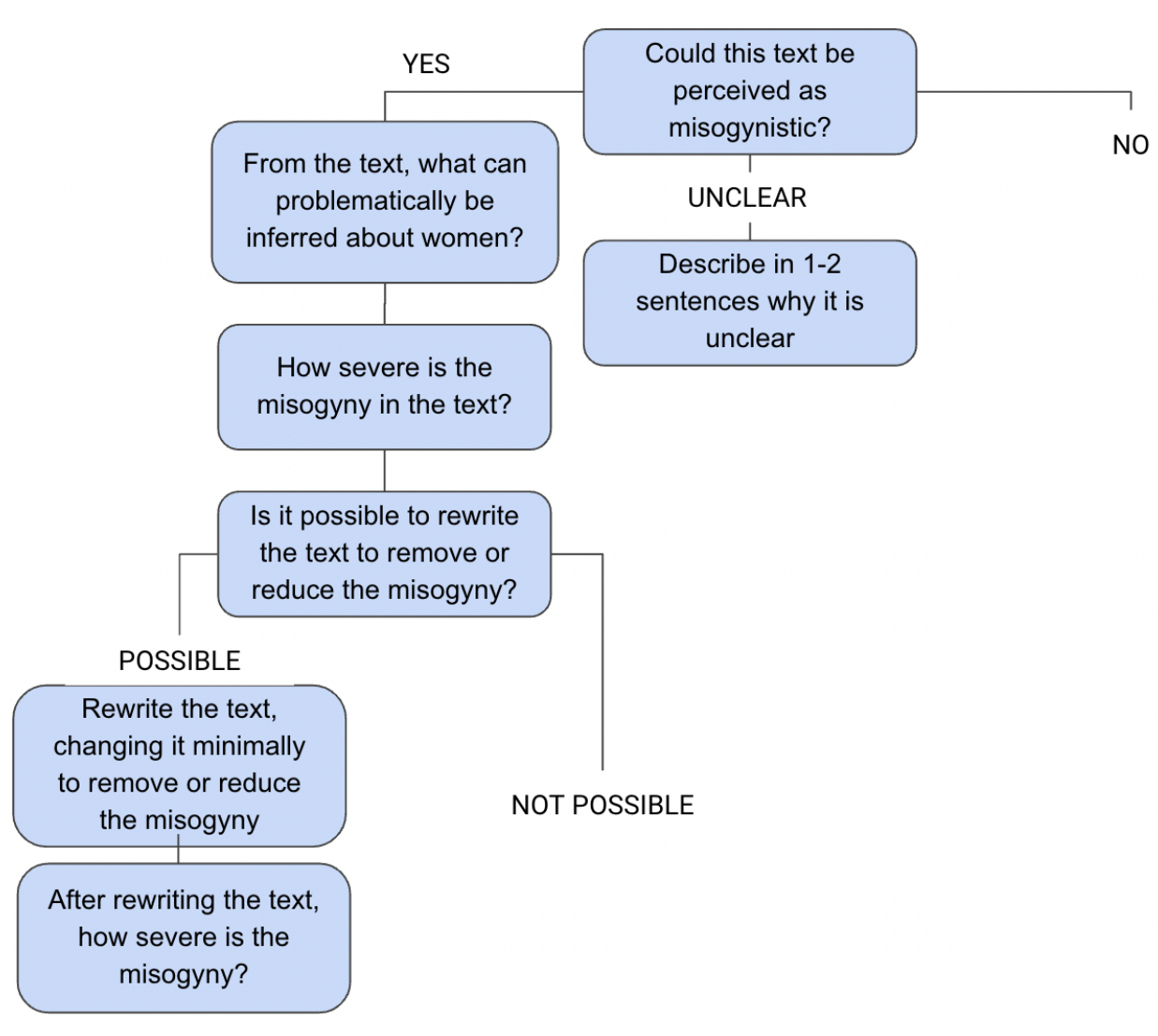}
\caption{Summary of the annotation tasks for the Biasly dataset.}
\label{fig1}
\end{figure}

All annotation tasks discussed in this section are summarized in Figure \ref{fig1}.

\textbf{Task 1:} Annotators were asked to conduct a binary classification (yes/no) of whether the data point presented contained misogyny anywhere within it. The annotators referenced the following definition of misogyny: “Hatred of, dislike of, contempt for, or ingrained prejudice against women." Misogyny may be directed at a group or an individual, but it is easier to detect in generalizations about groups. For individuals, there is additional verification necessary to be sure that the negative sentiment is, at least in part, associated with the individual in virtue of their being identified as a woman. We included reclaimed language, slurs, and potentially humorous utterances as misogynistic so as not to risk creating noise with a seemingly inconsistent dataset. While we recognize that some such speech does indeed have both misogynistic and non-misogynistic uses and that this would be an interesting subject of further study, our goal for this initial step was to identify what \textit{could be} misogynistic and not just what was definitely misogynistic in context, especially since our context was limited to three-sentence windows. Through consultation with annotators, it emerged that when female-oriented slurs were being described rather than used to exhibit misogynistic sentiments  (related to the use/mention distinction in linguistics), they did not find the descriptions to be misogynistic. As a result, while there may still be uses of other slurs where even mentioning them does evoke negative and potentially harmful sentiments (cf. \citet{davis2020instability}), for this corpus, we prescribed annotating descriptions as non-misogynistic and uses as misogynistic.

\paragraph{Task 2:} Once annotators identified a data point as containing misogyny, they were asked to classify the type(s) of misogyny being exhibited in the data point from a provided list that they had a hand in creating.
New categories were devised, ones that uniquely fit our dataset (i.e. gender essentialism and stereotypes, intersectional and identity-based misogyny, lacking autonomy/agency etc.). 

\begin{table*}[t]
    \centering
    \resizebox{\textwidth}{!}{%
    \begin{tabular}
    {|>{\RaggedRight}p{7cm}|m{11cm}|} \toprule 
         \textbf{Name} & \textbf{Description}\\
         \hline \midrule
         Anti-feminism& Feminism is a bad idea, feminists are gross and ugly, women shouldn’t have equal rights\\ \midrule
         Dehumanization& Comparing women to animals or objects\\ \midrule
         Domestic violence and other violence against women & (self-explanatory)\\ \midrule
         Gender essentialism or stereotypes & Can be both positive, e.g. women are good at childrearing and cooking because they are more nurturing, and negative, e.g. women are untrustworthy and overly emotional because of their hormonal cycles\\ \midrule 
         Gendered slurs & \textit{Chick}, \textit{b*tch}, \textit{c*nt}, etc.\\ \midrule
         Intersectional, identity-based misogyny&  Any other instance of misogyny that is related to race, ethnicity, religion, class, occupation, immigration status, disability, size, etc.\\ \midrule 
         Lacking autonomy or agency&  Women are not able to make decisions or must defer to male authorities\\ \midrule
         Phallocentrism & Focus on penis in organization of social world\\ \midrule 
         Rape and other forms of sexual violence&  (self-explanatory)\\ \midrule 
        Sexualization&Outsized focus on appearance, degrading language\\ \midrule 
        Transmisogyny/ Homophobia& Includes mocking individuals or groups for gender nonconformity, e.g. for dressing or acting in a way that does not conform with assumed gender roles; homophobia/transphobia that also contains misogynistic inferences\\
        \midrule
        Trivialization & Infantilizing or paternalistic language, women are not taken seriously\\
        \bottomrule
    \end{tabular}
    }
    \caption{Subcategories of misogyny with a short explanation}
    \label{tab:categories}
\end{table*}

Table \ref{tab:categories} shows the full list of the types for task 2. We established a misogyny categorization for two reasons. First, our annotators wanted a way to frame their thinking and move beyond their initial write-in categories in exploring each datapoint’s misogynistic inference(s). By categorizing misogynistic inferences using a pre-defined set, they could: i) think more strategically about other annotation tasks (i.e. how severe the misogyny is and whether it can be removed or lessened); and ii) have more overlap with other annotators who had gone through the same categorization exercise. Second, we wanted to be able to create models that help educate the end user about why their text might be considered misogynistic. An approach that focuses on education, rather than just correction, may be more conducive to understanding and behavioral change for end users, thereby enhancing the impact of this work \citep{malepsyc}.



\paragraph{Task 3:} In addition to categorizing misogynistic data points, annotators were asked to indicate the datapoint's severity on a continuous scale. The continuous scale (rather than ordinal) was intentionally chosen to acknowledge the impossibility of ascribing one definitive number to the severity of a misogynistic statement and to avoid the pitfall of using a discrete metric for a potentially continuous variable \citep{matejka2016effect}. The continuous scale allows for a more genuine reflection of human interpretations of misogyny. While annotators only saw the continuous scale with the endpoints of no misogyny/maximal misogyny, the back end was mapped to values between 0 and 1000.

\paragraph{Task 4:}
Last, annotators were asked, when handling misogynistic data points, whether it was possible to remove the misogynistic inference(s) by rewriting portions of the text while largely retaining the original meaning of the utterance(s). However, the feasibility of this task depended on whether a misogynistic inference was primary (the main point of the utterance) or secondary (e.g. an implicature). When it is primary, the rewrite task is likely impossible in the sense that annotators could not remove the misogyny without losing the core of the original sentence(s). When the misogynistic inference is secondary, the rewrite was more likely if there was a way to retain the primary intent of the speaker while removing the misogynistic inference.

\begin{table*}[t!]
\centering
\resizebox{\textwidth}{!}{%
\begin{tabular}{ l  |l  |m{8.5cm}| l|m{6cm} } \toprule 

\#& Sev.& Data Point & Category &Misogyny Mitigation \\ \midrule

1 & Low& She needs my support. Girl could you give us a second? Really? & Trivialization & She needs my support. Could you give us a second? Really? \\ \midrule

2 & Mid& I think it's about time that Emanuel had a nice fellow in her life. Why? Were you starting to think that I was a lesbian? & Stereotype &  NA \\ \midrule 

3 & High& We passed her mama around like a baton, man. Yeah. You never told me that about your mother. & Dehumanization & We all slept with her mama, man. Yeah. You never told me that about your mother. \\ \bottomrule

\end{tabular}
}
\caption{Example Annotations from the Biasly Dataset.}
\label{tab:examples}
\end{table*}

\subsection{Annotation settings}
\label{sec:annotation-settings}
\paragraph{Engaging Expert Annotators:}\label{annotators} 
When annotating for misogyny, a subjective, nuanced and political task, we wanted to ensure that the interpretation of each data point was grounded in research and expertise. As such, we hired annotators pursuing or having completed their post-secondary degrees in linguistics, gender studies, or both. We did not place other demographic limits on recruitment, and our annotators included a range of gender and sexual identities, races, ethnicities, and language backgrounds, though all were located in North America and were fluent in English.
 
While our annotators were using the annotation guidelines from which the above excerpts were drawn to guide their annotations of the final dataset, the guidelines were created in an iterative, collaborative manner: 
We held workshops and pilot rounds wherein annotators stress tested the initial version of the annotation tasks without strict prescriptive direction \citep{rottger}. Subsequently, we sought feedback through moderated discussions with the team's gender studies and linguistics experts, who crystallized our approach with prescriptive guidelines that were led by the annotators' comments. This grounded theory approach informed elements like our misogynistic inference categories, interpretations of severity, and appropriate rewrites \citep{grounded}. When devising the list of inference categories, for example, no categories were provided in a pilot round; annotators were asked to consider which misogynistic beliefs were being expressed in the data provided. In all cases, they brought their observations to a workshop guided by our experts who finalized the category list, severity, and rewrite guidelines based on existing literature, feedback, and their own analyses of the data. 

Our team was in close contact with the annotators throughout the process, hosting regular office hours and remaining available over Slack and email. Annotators connected amongst themselves via Slack, enabling them to discuss strategies for confusing or complex datapoints. We allowed space for real differences of opinion and did not suggest that a consensus was necessary. This close contact also allowed us to check in with the annotators about the potentially harmful impacts of working with misogynistic texts, which our annotators reported being able to manage well.

\paragraph{Inter-Annotator Distribution:}
Our team of annotators included 5 gender studies and 5 linguists experts (although expertise between the two groups at times overlapped). To check for inter-annotator agreement (and ensure quality control), three annotators were assigned to each data point. Gender studies and linguistics annotators were intentionally assigned at a 2:1 ratio to each data point to ensure a diversity of academic backgrounds were included in each annotation. We believe this allowed our dataset to contain strong linguistics and gender studies input.

\section{Dataset Analysis}

In Table \ref{tab:examples}, we have three examples with differing levels of severity of misogyny, from least to most. 

In each case, all three annotators agreed that it was misogynistic. (1) is an example of trivialization via infantilizing or paternalistic language (referring to a woman as a girl). (2) relies on gender essentialism or stereotypes (that women aren't complete without romantic attachments), and, as with many examples, is intersectional, combining misogyny with aspects of homophobia (or, in other cases, racism, ableism, etc.). Finally, (3) is dehumanizing in its comparison of a woman to an object. Some examples were deemed to be impossible to rewrite, as in (2) where all 3 annotators agreed that mitigation was not possible. Some were possible to rewrite with total mitigation (removal) of the misogynistic inference, as in (1), where the 3 annotators all provided the same rewrite, simply eliminating the problematic item with no significant effect on the dialogue. Finally, we have examples like (3) where rewriting is possible and can mitigate but not eliminate the misogyny. 

\paragraph{Quantitative Analysis:}
Our resulting dataset consists of 10000 datapoints that were each annotated 3 times, leading to a total set of 30000 annotations. 
5600 of the 30000 annotations were labelled as misogynist according to the first binary classification task (Task 1), with an inter-annotator agreement of 0.4722 according to Fleiss' kappa \citep{fleiss}. 
The severity of the original misogynist datapoints (from 0 to 1000) has a mean of 344.8 with a standard deviation of 209.1, while the severity of all rewritten datapoints has a mean of 53.6 and a standard deviation of 115.8, reflecting a significant reduction in misogyny severity.  
The most frequent sub-category of misogyny was \textit{Trivialization} with 2227 occurrences, while \textit{Transmisogyny} only appeared 43 times.
1985 misogynist datapoints were selected to be rewritten to mitigate the misogyny by one or more annotators, yielding a total of 2977 rewrites.

In order to perform binary classification of misogyny with the annotated dataset, we needed to relate each datapoint to only one label. Since the focus of our dataset is to identify subtle forms of misogyny, we aggregated the binary classifications from all annotators into a single label by deeming the data point misogynistic as soon as one of the three annotators labels it so. This way, we ensure that we are capturing even the subtlest forms of misogyny and prevent overriding minority voices with a `majority rules’ approach.
Using this methodology, our dataset contains 3159 misogynist datapoints, which gives a distribution of 31.59\% positive cases and 68.41\% negative cases, a more balanced distribution than much previous work in the field has achieved \citep{Edos,Guest,callme}.

\section{Baseline Results}
This section describes the experimental setups for machine learning models on our dataset and presents the results. To follow best practice from other work in the field \citep{Edos,Guest,Ami,callme}, we provide baseline results for the machine learning tasks of misogyny classification, severity regression and mitigation by rewriting, exemplifying a few tasks for which our dataset can be used, while leaving multi-label classification for future work.

\subsection{Experimental Setups}

\paragraph{Classification:}
 For our classification experiments, we used four models and report the F1 scores: BERT \citep{bert}, RoBERTa \citep{roberta}, DeBERTa v3 \citep{deberta}, and ELECTRA \citep{electra}. For each, we used a train/eval/test split of 80/10/10.
In all four cases, we used the base version with a maximum input sequence length of 512, batch size of 32, a learning rate of 2e-5, and 3 epochs for training. 

\paragraph{Severity:}
We fine-tuned a BERT regression model to predict the misogyny severity scores (Task 3) in a supervised manner.
Following \citet{callme}, we additionally report the (unsupervised) Perspective API \citep{perspectiveapi} toxicity scores for our data. For the regression experiment, as well as to compare the severity to the Perspective API toxicity scores, the original severity values were transformed from a range of [0,1000] to [0,1].
We used 80 percent of the dataset for training (with 10 percent allocated for evaluation) and 20 percent for testing. To ensure comprehensive coverage, we included non-misogynistic data points with a severity score of 0. For data points labeled as misogynistic by at least one annotator, we computed the average of severity scores provided by all annotators who identified them as such. We fine-tuned a BERT (bert-base-uncased) model for linear regression over three epochs, with a learning rate of 2e-5, a weight decay of 0.1, and a per-device train batch size of 64.

\paragraph{Mitigation:} For misogyny mitigation (Task 4), we fine-tuned three baseline models: BART \cite{lewis-etal-2020-bart}, FLAN-T5 \cite{chung2022scaling}, and Alpaca-LoRA \cite{alpaca-lora}.
Following the methodology outlined in the ParaDetox paper, all our experiments across various models adhered to specific hyperparameters, including a learning rate of 3e-5, a total of 100 training epochs, and a gradient accumulation step of 1. Additionally, we employed the `base' version of each model. During training, we conducted evaluations after each epoch and selected the checkpoint with the lowest loss on the evaluation set for subsequent prediction tasks. 


\begin{table*}[t!]
\resizebox{\textwidth}{!}{%
\begin{tabular}{l|l|cc|ccc|ccc}
model & dataset & accuracy$\uparrow$ & F1\_macro$\uparrow$  & precision\_yes$\uparrow$  & recall\_yes$\uparrow$  & F1\_yes$\uparrow$  &precision\_no$\uparrow$  & recall\_no$\uparrow$  & F1\_no$\uparrow$ \\
 \toprule
BERT & Edos & 0.874 & 0.826 & 0.753 & 0.718 & 0.735 & 0.911 & 0.925 & 0.918 \\
 & Guest & 0.937 & 0.806 & 0.728 & 0.581 & 0.647 & 0.955 & 0.976 & 0.965 \\
 & AmiData & 0.691 & 0.691 & 0.631 & 0.791 & 0.702 & 0.773 & 0.606 & 0.679 \\
 & Callme & 0.929 & 0.865 & 0.738 & 0.809 & 0.772 & 0.966 & 0.950 & 0.958 \\
 & Ours  & 0.829 & 0.801 & 0.738 & 0.717 & 0.727 & 0.870 & 0.881 & 0.875 \\
\midrule
DEBERTA & Edos & 0.873 & 0.824 & 0.751 & 0.713 & 0.732 & 0.910 & 0.924 & 0.917 \\
 & Guest & 0.938 & 0.803 & 0.750 & 0.558 & 0.640 & 0.953 & 0.980 & 0.966 \\
 & AmiData & 0.709 & 0.707 & 0.635 & 0.863 & 0.732 & 0.832 & 0.578 & 0.682 \\
 & Callme & 0.936 & 0.877 & 0.757 & 0.830 & 0.792 & 0.970 & 0.954 & 0.962 \\
 & Ours  & 0.817 & 0.790 & 0.710 & 0.717 & 0.714 & 0.867 & 0.864 & 0.866 \\
\midrule
ELECTRA & Edos & 0.877 & 0.830 & 0.763 & 0.718 & 0.740 & 0.911 & 0.929 & 0.920 \\
 & Guest & 0.924 & 0.765 & 0.647 & 0.512 & 0.571 & 0.947 & 0.969 & 0.958 \\
 & AmiData & 0.705 & 0.704 & 0.641 & 0.817 & 0.718 & 0.797 & 0.609 & 0.690 \\
 & Callme & 0.930 & 0.871 & 0.720 & 0.860 & 0.784 & 0.975 & 0.942 & 0.958 \\
 & Ours & 0.819 & 0.791 & 0.716 & 0.714 & 0.715 & 0.867 & 0.868 & 0.867 \\
\midrule
ROBERTA & Edos& 0.878 & 0.834 & 0.747 & 0.749 & 0.748 & 0.920 & 0.919 & 0.919 \\
 & Guest& 0.929 & 0.786 & 0.673 & 0.558 & 0.610 & 0.952 & 0.970 & 0.961 \\
 & AmiData & 0.712 & 0.712 & 0.669 & 0.739 & 0.702 & 0.756 & 0.689 & 0.721 \\
 & Callme & 0.937 & 0.880 & 0.763 & 0.836 & 0.798 & 0.971 & 0.955 & 0.963 \\
 & Ours & 0.816 & 0.788 & 0.709 & 0.714 & 0.712 & 0.866 & 0.864 & 0.865
\end{tabular}
}
\caption{Test results of various binary classification models on our dataset and on other misogyny classification datasets we mention in this paper.}
\label{table:results_classification}
\end{table*}

\subsection{Results}

\subsubsection{Classification}
\label{sec:further-results-classification}

Following related work \citep{Edos, callme, Guest, Ami}, we use the macro-F1 score for our evaluations to account for the class imbalance between misogynistic and non-misogynistic datapoints.
We provide the results on the test set of the four models BERT \citep{bert}, DeBERTa \citep{deberta}, ELECTRA \citep{electra}, and RoBERTa \citep{roberta}, all fine-tuned on our dataset in Table 4.
We additionally provide the results for the same models with identical setup, fine-tuned on the respective training set and tested on the respective test set of each, for the EDOS dataset \citep{Edos}, the dataset from \citep{Guest}, the Ami dataset \citep{Ami} and the Callme dataset \citep{callme}, as we included those in the comparison in Table \ref{tab:dataset_comparison}.

\subsubsection{Severity}
\label{sec:app-regression}
For the severity score, we provide performance results on the test set for the fine-tuned BERT and also use regression metrics to compare the Perspective API toxicity scores to the severity scores of our annotators in Table 5.
We selected four principal metrics for evaluation. First, Mean Absolute Error (MAE) elucidates the average magnitude of errors, offering an easily interpretable representation of the proximity of predictions to actual values. Mean Squared Error (MSE) and Root Mean Squared Error (RMSE) provide insights into the performance of the model by minimizing squared errors, with RMSE presenting an error metric in the same unit as the target variable; in our case, the severity scores in a (transformed) range of [0,1]. Lastly, the Coefficient of Determination or R-squared (R²) delineates the extent of the variance in the target variable explained by the model.

\begin{table}[h]

\centering
\resizebox{\columnwidth}{!}{%
\begin{tabular}{l|lllll}
 & mse$\downarrow$ & rmse$\downarrow$ & mae$\downarrow$ & r2$\uparrow$ \\
 \toprule

perspective\_toxicity & 0.078 & 0.280 & 0.191 & -1.761 \\

\midrule
BERT\_test & \textbf{0.020} & \textbf{0.143} & \textbf{0.098} & \textbf{0.324} \\

\end{tabular}
}
\caption{Test results of supervised and unsupervised toxic regression models on misogyny regression. We can see that BERT trained with supervised learning performs better for predicting the level of misogyny as compared to the Perspective AI toxicity score.}

\end{table}

\subsubsection{Mitigation}
\label{rewrite_results}

We report metrics for the rewrite task in Table 6. The BLEU metric uses the BLEU score compared to the human reference. We evaluate Content Preservation (SIM) using the cosine similarity between the embeddings of the original text and the output, computed utilizing the model described in \citet{wieting-etal-2019-beyond}. The Style Accuracy (STA) metric represents the percentage of non-toxic outputs as identified by a style classifier, as detailed in \citet{logacheva-etal-2022-paradetox}. For toxicity scores and sexually explicit toxicity scores (Gen. toxicity), we use Perspective API to obtain the scores and compared them to those of the inputs and references.

\begin{table*}[h]
\resizebox{\textwidth}{!}{%
\begin{tabular}{l|l|rrr|rrr|rrr}
 &  & BLEU$\uparrow$ & SIM$\uparrow$ & STA$\uparrow$ & Toxicity& Tox-Inp & Tox-Refs & Gen. Tox. & Gen.-Inp & Gen.-Refs \\
 \toprule
BART & ParaDetox & 56.00 & 0.87 & 0.86 & 18.57 & -55.61 & 4.22 & 2.71 & -16.4 & 1.00 \\
 & Appdia & 58.8 & 0.92 & 0.58 & 41.12 & -25.08 & 18.75 & 7.74 & -9.74 & 5.31 \\
 & Ours & 85.77 & 0.96 & 0.75 & 29.46 & -4.89 & 3.56 & 13.46 & -2.62 & 2.64 \\
 \midrule
FLAN-T5 & ParaDetox & 53.43 & 0.87 & 0.88 & 17.53 & -56.65 & 3.18 & 2.12 & -16.99 & 0.41 \\
 & Appdia & 57.21 & 0.87 & 0.72 & 33.69 & -32.5 & 11.33 & 4.51 & -12.97 & 2.08 \\
 & Ours & 86.94 & 0.97 & 0.75 & 29.84 & -4.51 & 3.94 & 13.41 & -2.67 & 2.59 \\
 \midrule
Alpaca-LoRA & ParaDetox & 55.97 & 0.89 & 0.8 & 22.08 & -52.1 & 7.74 & 3.61 & -15.5 & 1.90 \\
 & Appdia & 60.48 & 0.83 & 0.73 & 22.08 & -52.1 & 7.73 & 3.61 & -15.5 & 1.90 \\
 & Ours & 86.36 & 0.95 & 0.76 & 27.52 & -6.83 & 1.62 & 11.91 & -4.17 & 1.09
\end{tabular}
}
\caption{Test results of different text generator models for misogyny mitigation.}
\end{table*}

Upon comparing results across various rewrite datasets, it is apparent that our dataset predominantly focuses on misogynistic rewrites, more so than either the ParaDetox or Appdia dataset. Even though different models attain high BLEU scores relative to the references—thanks to our annotators who are instructed to alter the text minimally—the gender toxicity levels in the rewrites remain notably high in our dataset. This observation underscores that our dataset is well-suited for examining misogynistic rewrites. It is important to note that, based on Perspective API scores, the models are far from significantly reducing sexual explicitness. This outcome predominantly stems from our deliberate choice to prioritize the preservation of semantic meaning over extensive content alteration, a strategy that differentiates our approach from those adopted in creating the other two datasets mentioned.

\section{Conclusion and Future Work}
Developed through collaboration among domain experts in gender studies, linguistics, and natural language processing, we present Biasly, a comprehensive dataset designed for misogyny detection, regression analysis, and mitigation. We have carefully collected 10,000 annotations for classification. For those classified as misogynistic by the annotators, the corpus provides rewrites
where possible and misogyny severity scores before and after the rewrite.

In future work, we aim to employ more advanced modeling techniques, such as in \citet{davani-etal-2022-dealing}, to incorporate diverse annotator perspectives, moving beyond our current reliance on a single label for the baselines presented here. We also plan to provide baselines for the multi-label classification task of categorizing misogynistic inferences and to release the dataset itself as an open-source resource along with its datasheet and bias statement. Our hope is that Biasly serves as a model for socially responsible dataset creation for language models. This process can be readily applied to diverse domains, fostering a broader commitment to responsible AI development.

\section{Social Impacts Statement}
While we see many beneficial applications of this dataset, namely in building future applications designed to educate the public about misogyny, how it is expressed, and ways it can be removed or minimized, this dataset also presents the risk of being used for nefarious purposes. Specifically, malicious actors could use the dataset to create content that evades traditional toxicity detection models by rendering the misogynistic text more subtle. Furthermore, one could leverage this model to introduce subtle bias into otherwise non-misogynistic statements. This is one of the reasons we're looking to support the development of tools for more robust detection of misogyny, which can identify misogyny in subtle forms as well as overt. In other words, part of our desire to contribute to the domain of subtle misogyny detection is so that this type of misogyny doesn't continue to go unnoticed by traditional toxicity detection tools. 


From a development standpoint, the risks centered mostly around our annotators, specifically in terms of their repeat exposure to misogynistic content, particularly data points which mentioned violence or suicide. In order to protect our annotators as much as possible, we shared mental health resources accessible through their respective universities, conducted mental health check-ins through surveys, and provided an opportunity to meet with members of our team to discuss the impact any of the work was having on their mental health. Furthermore, our team was easily accessible through platforms that allowed for direct communication. Overall, 70 percent of our annotators said they were fairly comfortable with the task in the context of our project (four people ranked their comfort at 4 out of 5 and three people ranked their comfort at 5 out of 5 in a survey). We made sure to address the feedback we had received in the free text portions of our survey to accommodate the needs expressed (i.e. offering to meet with annotators one-on-one to discuss material they find distressing). We’d like to continue treating annotators as key team members in the project and plan on hosting information sessions to share the impact of their contributions.

\bibliography{custom,anthology}

\newpage
\appendix
\section{Appendix}

\subsection{Annotation Interface}
\label{app-interface}

A screen capture of the annotation interface provided to  our annotators is given in Figure \ref{fig:interface}.

\noindent\begin{minipage}{\textwidth}
    \centering
\vspace{1cm}
\includegraphics[width=\textwidth]{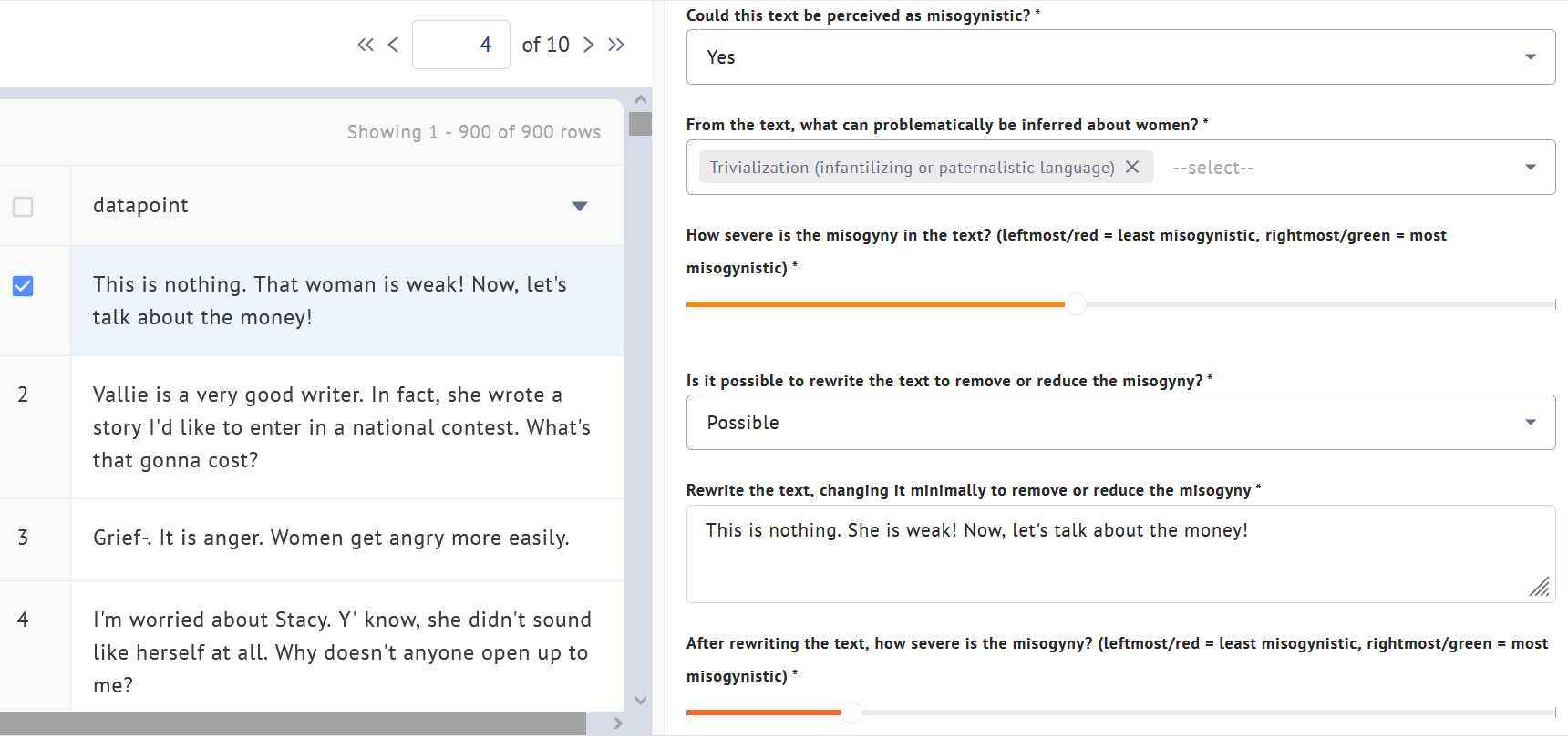}
\captionof{figure}{Screen capture of the annotation interface.}
\label{fig:interface}

\end{minipage}




\end{document}